\documentclass[11pt]{article}
\usepackage{amssymb}
\usepackage{epsfig}
\usepackage{fullname}
\usepackage{fullpage}
\newcommand{\epsfscaledbox}[2]{\centerline{\psfig{figure=#1,width=#2}}}
\newcommand{\omt}[1]{}

\newcommand{\nSet}{N}
\newcommand{\set}[1]{\mbox{$\{#1\}$}}
\newcommand{\igt}{I_{>}}
\newcommand{\vote}[2]{v_{#1}(#2)}
\newcommand{\totalvote}[1]{v_{\nSet}(#1)}
\newcommand{\gloss}[1]{{\sf #1}}
\newcommand{\D}{{\rm D}}
\newcommand{\W}{{\rm W}}
\newcommand{\C}{{\rm C}}
\newcommand{\X}{{\rm X}}
\newcommand{\sst}{{\sc sst}}
\newcommand{\ours}{{\sc tango}}
\newcommand{\hpt}{.}

\begin{document}

\title{\vspace{-75pt}
{\normalsize \tt \hfill Preliminary version; to appear in {\em Natural Language Engineering}, 2003} \\ \mbox{}\\
Mostly-Unsupervised Statistical Segmentation of Japanese
Kanji Sequences}
\author{
\begin{tabular}{c@{~~~~~~}c}
Rie Kubota Ando & Lillian Lee \\
IBM Thomas J. Watson Research Center & Dept. of Computer Science \\
P.O. Box 704& Cornell University \\
Yorktown Heights, NY 10598, USA & Ithaca, NY 14853 USA \\
{\tt rie1@us.ibm.com} & {\tt llee@cs.cornell.edu}
\end{tabular}
}

\maketitle

\begin{abstract}
Given the lack of word delimiters in written Japanese, word
segmentation is generally considered a crucial first step in
processing Japanese texts.  Typical Japanese segmentation algorithms
rely either on a lexicon and syntactic analysis or on pre-segmented
data; but these are labor-intensive, and the lexico-syntactic
techniques are vulnerable to the {\em unknown word problem}.  In
contrast, we introduce a novel, more robust statistical method
utilizing {\em unsegmented} training data. Despite its simplicity, the
algorithm yields performance on long kanji sequences comparable to and
sometimes surpassing that of state-of-the-art morphological analyzers
over a variety of error metrics.  The algorithm also outperforms
another mostly-unsupervised statistical algorithm previously proposed
for Chinese.

Additionally, we present a two-level annotation scheme for Japanese to
incorporate multiple segmentation granularities, and introduce two
novel evaluation metrics, both based on the notion of a {\em
compatible bracket}, that can account for multiple granularities
simultaneously.

\end{abstract}

\section{Introduction}

Because Japanese is written without delimiters between words (the
analogous situation in English would be if words were written without
spaces between them), accurate {\em word segmentation} to recover the
lexical items is a key first step in Japanese text
processing. Furthermore, word segmentation can also be used as a more
directly enabling technology in applications such as extracting new
technical terms, indexing documents for information retrieval, and
correcting optical character recognition (OCR) errors
\cite{Wu+Tseng:93a,Nagao+Mori:94a,Nagata:96a,Nagata:96b,Sproat+al:96a,Fung:98a}.

Typically, Japanese word segmentation is performed by morphological
analysis based on lexical and syntactic knowledge.  However,
character sequences consisting solely of {\em kanji} (Chinese-derived
characters, as opposed to hiragana or katakana, the other two types of
Japanese characters) pose a major challenge to morphologically-based
segmenters for several reasons.

First and most importantly, lexico-syntactic morphological analyzers
are vulnerable to the {\em unknown word problem}, where terms from the
domain at hand are not in the lexicon; hence, these analyzers are not
robust across different domains, unless effort is invested to modify
their databases accordingly.  The problem with kanji sequences is
that they often contain domain terms and proper nouns, which are
likely to be absent from general lexicons: Fung \shortcite{Fung:98a}
notes that 50-85\% of the terms in various technical dictionaries are
composed at least partly of kanji.  Domain-dependent terms are quite
important for information retrieval, information extraction, and text
summarization, among other applications, so the development of
segmentation algorithms that can efficiently adapt to different
domains has the potential to make substantial impact.

Another reason that kanji sequences are particularly challenging for
morphological analyzers is that they often consist of
compound nouns, so syntactic constraints are not applicable.  For
instance, the sequence
\psfig{figure=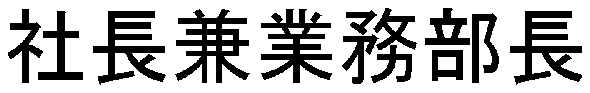,width=2.cm}, whose
proper segmentation is
\psfig{figure=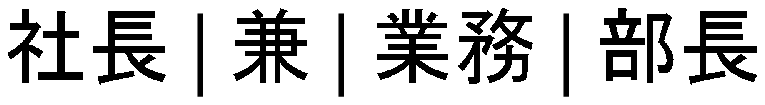,width=2.8cm}
(\gloss{president $\vert$ both-of-them $\vert$ business $\vert$ general
manager}, i.e., \gloss{a president as well as a general manager of
business}), could be incorrectly segmented
as \psfig{figure=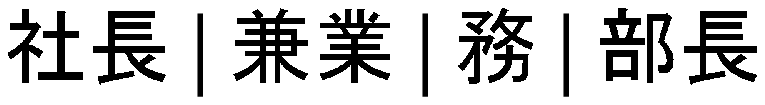,width=2.8cm}(\gloss{president
$\vert$ subsidiary business $\vert$ (the name) Tsutomu $\vert$ general
manager}); since both alternatives are four-noun sequences, they
cannot be distinguished by part-of-speech information alone.

An additional reason that accuracy on kanji sequences is an
important aspect of the total segmentation process is that they
comprise a significant portion of Japanese text, as shown in Figure
\ref{fig:corpusstats}.
\begin{figure}[t]
\begin{center}                    
\begin{tabular}{lrr}
Sequence length  &   \# of characters  & \% of corpus \\ \hline

       1 - 3 kanji      &  20,405,486     &      25.6 \\
       4 - 6 kanji      &  12,743,177    &      16.1   \\
       more than 6 kanji &  3,966,408    &       5.1  \\ \hline
       Total                 &  37,115,071    &      46.8  \\
\end{tabular}
\end{center}
\caption{\label{fig:corpusstats} Kanji statistics from 1993 Japanese
newswire (NIKKEI).  The sequence length categories are disjoint; to
take an analogous example from English: ``theQUICKbrownFOX" has three
lower-case characters in sequences of length 1-3 and five lower-case
characters in sequences of length 4-6.}
\end{figure}
Since sequences of more than 3 kanji generally consist of more than
one word,  at least 21.2\% of 1993 Nikkei newswire
consists of kanji sequences requiring segmentation; and long sequences
are the hardest to segment \cite{Takeda+Fujisaki:87a}.

As an alternative to lexico-syntactic and supervised
approaches,
we propose a simple, efficient segmentation method,
the \ours\ algorithm,
 which learns mostly from
very large amounts of {\em unsegmented} training data, thus avoiding
the costs of 
encoding lexical or syntactic knowledge or hand-segmenting large
amounts of training data.  Some key advantages of 
\ours\ are:
\begin{itemize}
\item A very small number of pre-segmented training examples (as few
as 5 in our experiments) are needed for good performance.
\item For long kanji strings, the method produces results rivalling
those produced by Juman 3.61 \cite{Kurohashi+Nagao:99a} and Chasen 1.0
\cite{Matsumoto+al:97a}, two morphological analyzers in widespread
use.  For instance, we achieve on average 5\% higher {\em word
precision} and 6\% better {\em morpheme recall}.
\item No domain-specific or even Japanese-specific rules are employed, enhancing
portability to other tasks and applications.
\end{itemize}

We stress that we explicitly focus on long kanji sequences, not only
because they are important, but precisely because traditional
knowledge-based techniques are expected to do poorly on them.  We view
our mostly-unsupervised algorithm as {\em complementing}
lexico-syntactic techniques, and envision in the future a hybrid
system in which our method would be applied to long kanji sequences
and morphological analyzers would be applied to other sequences for
which they are more suitable, so that the strengths of both types of
methods can be integrated effectively.

\subsection{Paper organization} This paper is organized as follows.  Section \ref{sec:alg} describes
our 
mostly-unsupervised, knowledge-lean 
algorithm.  Section \ref{sec:baselines} describes the morphological
analyzers Juman and
Chasen;
comparison against these methods, which rely on large lexicons and
syntactic information,  forms the
primary focus of our experiments.

In Section \ref{sec:exp-framework}, we give the details of our
experimental framework, including the evaluation metrics we
introduced.  Section \ref{sec:results} reports the results of our
experiments. In particular, we see in Section \ref{sec:cf-morph} that
the performance of our statistics-based algorithm rivals that of Juman
and Chasen.  In order to demonstrate the usefulness of the particular
simple statistics we used, in Section \ref{sec:cf-sun} we show that
our algorithm substantially outperforms a method based on {\em mutual
information} and {\em t-score}, sophisticated statistical functions
commonly used in corpus-based natural language processing.

We discuss related work in section
\ref{sec:related}, and conclude in section \ref{sec:conc}.

\section{The TANGO Algorithm}
\label{sec:alg}

\begin{figure}

\begin{center}
\epsfscaledbox{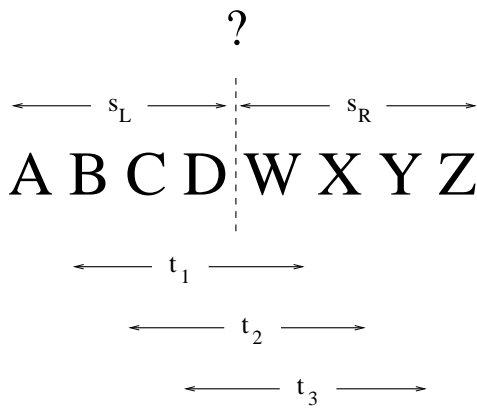}{2.5in}
\end{center}
\caption{\label{fig:intuition} Collecting 4-gram evidence for a word boundary
-- are the non-straddling $n$-grams $s_L$ and $s_R$ more frequent than
the straddling $n$-grams $t_1, t_2,$ and $t_3$?}
\end{figure}

Our algorithm, \ours\ ({\bf T}hreshold {\bf A}nd maximum for {\bf
N}-{\bf G}rams that {\bf
O}verlap\footnote{Also,\psfig{figure=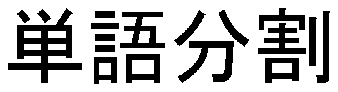,width=1.3cm}
({\em tan-go bun-katsu}) is Japanese for ``word segmentation''.}),
employs character $n$-gram counts drawn from an unsegmented corpus to make
segmentation decisions.  We start by illustrating the underlying
motivations, using the example situation depicted in
Figure \ref{fig:intuition}: ``A B C D W X Y Z'' represents a
sequence of  eight kanji characters, and the goal is to determine
whether there should be a word boundary between D and W.

The main idea is to check whether $n$-grams that are adjacent to the
proposed boundary, such as the 4-grams $s_L = $``A B C D'' and $s_R =
$``W X Y Z'', tend to be more frequent than $n$-grams that straddle
it, such as the 4-gram $t_1 =$ ``B C D W''.  If so, we have evidence
of a word boundary between D and W, since there seems to be relatively
little cohesion between the characters on opposite sides of this gap.

The $n$-gram orders used as evidence in the segmentation decision are
specified by the set $\nSet$.  For instance, our example in Figure
\ref{fig:intuition} shows a situation where $\nSet = \set{4}$,
indicating that only  4-grams are to be used. We thus pose the six
questions of the form, ``Is $\#(s_d) > \#(t_j)$?'', where $\#(x)$
denotes the number of occurrences of $x$ in the (unsegmented) training
corpus, $d \in \set{L,R}$, and $j \in \set{1,2,3}$.  If $\nSet =
\set{2,4}$, then two more questions (Is ``$\#(\mbox{C D}) > \#(\mbox{D
W})$?'' and ``Is $\#(\mbox{W X}) > \#(\mbox{D W})$?'') are used as
well. Each affirmative answer makes it more reasonable to place a
segment boundary at the location under consideration.

More formally, fix a location $k$ and an $n$-gram order $n$, and let
$s^n_L$ and $s^n_R$ be the non-straddling $n$-grams just to the left
and right of it, respectively. For $j \in \set{1,2,\ldots,n-1}$, let
$t^n_j$ be the straddling $n$-gram with $j$ characters to the right of
location $k$. We define $\igt(y,z)$ to be the indicator function that is 1 when
$y > z$, and 0 otherwise --- this function formalizes the notion of
``question'' introduced above.

\ours\ works as follows. It first calculates the fraction of
affirmative answers separately for each $n$-gram order $n$ in $\nSet$,
thus yielding a ``vote'' for that order $n$:
$$ \vote{n}{k} = \frac{1}{2(n-1)} \sum_{d\in\set{L,R}} \sum_{j=1}^{n-1}
I_{>}(\#(s^n_d), \#(t^n_j)).$$
Note that in locations near the beginning or end of a sequence, not
all the $n$-grams may exist (e.g., in Figure \ref{fig:intuition},
there is no non-straddling 4-gram to the left of the A-B gap), in
which case we only rely on comparisons between the existing relevant
$n$-grams --- since we are dealing with long kanji sequences, each
location will have at least one $n$-gram adjacent to it
and one $n$-gram straddling it, for reasonable values of $n$.

Then, \ours\ averages together the 
votes of each $n$-gram order:
$$\totalvote{k} = \frac{1}{|\nSet|}\sum_{n \in \nSet}\vote{n}{k}~ . $$
Hence, $\totalvote{k}$, or ``total vote'',  represents the average amount of evidence,
according to the participating $n$-gram lengths, that there should be a word
boundary placed at location $k$.

\begin{figure}
\epsfscaledbox{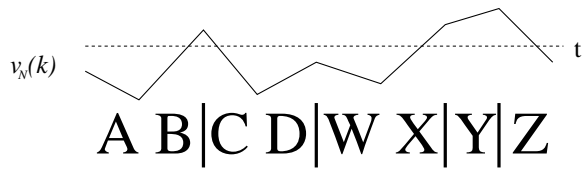}{3in}
\caption{\label{fig:splitdec} 
Determining word boundaries.  By the
threshold criterion, a boundary should be placed
at the B-C, X-Y, and Y-Z gaps;  by the local maximum criterion, a
boundary should be placed at the B-C, D-W, and Y-Z gaps.  Hence, four
boundaries are placed as shown.}
\end{figure}

After $\totalvote{k}$ is computed for every location,
boundaries are placed at all locations $\ell$ such that either:
\begin{enumerate}
\item $\totalvote{\ell} > \totalvote{\ell-1}$ and $\totalvote{\ell} >
\totalvote{\ell+1}$,  or 
\item $\totalvote{\ell} \geq t$, a threshold parameter.  
\end{enumerate}
That is, a segment boundary is created if and only if 
the total vote
is either (1)  a local
maximum, or (2) exceeds threshold $t$ (see Figure
\ref{fig:splitdec}).  The second condition is necessary for the
creation of single-character words, since local maxima by definition
cannot be adjacent. Note that the use of a threshold parameter also
allows a degree of control over the granularity of the segmentation:
low thresholds encourage shorter segments.

One may wonder why we only consider comparisons between $n$-grams of
the same length.  The underlying reason is that counts for longer
$n$-grams can be more unreliable as evidence of true frequency; that
is, the sparse data problem is worse for higher-order $n$-grams.
Because they are potentially unreliable, we do not want longer
$n$-grams to dominate the segmentation decision; but since there are
more overlapping higher-order $n$-grams than lower-order ones, we
compensate by treating the different orders separately.  Furthermore,
counts for longer $n$-grams tend to be smaller than for shorter ones,
so it is not clear that directly comparing counts for different-order
$n$-grams is meaningful. One could use some sort of normalization
technique, but one of our motivations was to come up with as simple an
algorithm as possible.\footnote{While simplicity is itself a virtue,
an additional benefit to it is that having fewer variables reduces the
need for parameter-training data.}

While of course we do not make any claims that this method has any
psychological validity, it is interesting to note that recent studies
\cite{Saffran+Aslin+Newport:96a,Saffran:01a} have shown that infants
can learn to segment words in fluent speech based only on the
statistics of neighboring syllables.  Such research lends credence to
the idea that simple co-occurrence statistics contain much useful
information for our task.

\subsection{Implementation Issues}

In terms of implementation, both the count acquisition phase (which
can be done off-line) and the testing phase are efficient.  Computing
$n$-gram statistics for all possible values of $n$ simultaneously can
be done in $O(m \log m)$ time
using suffix arrays, where $m$ is the training corpus size
\cite{Manber+Myers:93a,Nagao+Mori:94a,Yamamoto+Church:01a}.\footnote{Nagao
and Mori \shortcite{Nagao+Mori:94a}
and Yamamoto and Church \shortcite{Yamamoto+Church:01a} use suffix arrays in the service of
unsupervised term extraction, which is somewhat related to the
segmentation problem we consider here.}
  However, if the set $\nSet$
of $n$-gram orders is known in advance, conceptually simpler
algorithms suffice.  For example, we implemented an $O(m \log m)$
procedure that explicitly stores the $n$-gram counts in a table.
(Memory allocation for count tables can be significantly reduced by
omitting $n$-grams occurring only once and assuming the count of
unseen $n$-grams to be one.  For instance, the kanji $n$-gram table
for $\nSet=\set{2,3,4,5,6}$ extracted from our 150-megabyte corpus was
18 megabytes in size, which fits easily in memory.)
In the application phase, \ours's running time is clearly linear in
the test corpus size if $|\nSet|$ is treated as a constant, since the
algorithm is quite simple.

We observe that some pre-segmented data is necessary in order to
set the parameters $\nSet$ and $t$. However, as described below,  very
little such data was required to get good performance; we therefore
deem our algorithm to be ``mostly unsupervised''.

\section{Morphological Analyzers: Chasen and Juman}
\label{sec:compete}
\label{sec:baselines}

In order to test the effectiveness of \ours, we compared it against
two different types of methods, described below.  The first type, and
our major ``competitor'', is the class of morphological analyzers,
represented by Chasen and Juman -- our primary interest is in whether
the use of large amounts of unsupervised data can yield information
comparable to that of human-developed grammars and lexicons. 
(We also compared \ours\ against the mostly-unsupervised statistical
algorithm developed by Sun et al. \shortcite{Sun+Shen+Tsou:98a},
described in Section \ref{sec:sun}.)

Chasen 1.0\footnote{
http://cactus.aist-nara.ac.jp/\-lab/\-nlt/\-chasen.html}
\cite{Matsumoto+al:97a} and Juman 3.61\footnote{
http://pine.kuee.kyoto-u.ac.jp/\-nl-resource/\-juman-e.html}
\cite{Kurohashi+Nagao:99a} 
are two state-of-the-art, publically-available, user-extensible systems.
In our experiments, the grammars of both systems were used as
distributed (indeed, they are not particularly easy to make additions
to; creating such resources is a difficult task).  The sizes of
Chasen's and Juman's default lexicons are approximately 115,000 and
231,000 words, respectively.

An important question that arose in designing our experiments was how
to enable morphological analyzers to make use of the parameter-training
data that \ours\ had access to, since the analyzers do not have parameters to tune. The only significant
way that they can be updated is by changing their grammars or
lexicons, which is quite tedious (for instance, we had to add
part-of-speech information to new entries by hand). We took what we
felt to be a reasonable, but not too time-consuming, course of
creating new lexical entries for all the bracketed words in the
parameter-training data (see Section \ref{sec:annotate}).  Evidence
that this was appropriate comes from the fact that these additions
never degraded test set performance, and indeed improved it  in some
cases.
Furthermore,
our experiments showed that reducing the amount of training
data hurt performance (see Section \ref{sec:analysis}), indicating
that the information we gave the analyzers from the segmented training
data is useful.

It is important to note that in the end, we are comparing algorithms
with access to different sources of knowledge.  Juman and Chasen use
lexicons and grammars developed by human experts. Our algorithm, not
having access to such pre-compiled knowledge bases, must of necessity
draw on other information sources (in this case, a very large
unsegmented corpus and a few pre-segmented examples) to compensate for
this lack. Since we are interested in whether using simple statistics
can match the performance of labor-intensive methods, we do not view
these information sources as conveying an unfair advantage, especially
since the annotated training sets were small, available to the
morphological analyzers, and disjoint from the test sets.

\section{Experimental Framework}
\label{sec:exp-framework}

In this section, we describe our experimental setup.
Section \ref{sec:data} gives details on the data we used 
and how
training proceeded.  Section
\ref{sec:annotate} presents the annotation policies we followed to segment
the test and parameter-training data.  Section \ref{sec:eval-measures}
describes the evaluation measures we used; these include fairly
standard measures, such as precision and recall, and novel measures,
based on the notion of {\em compatible brackets}, that  we
developed to overcome some of the shortcomings of the standard
measures in the case of evaluating segmentations.

\subsection{Data and Training Procedures}
\label{sec:data}
Our experimental data was drawn from 150 megabytes of 1993 Nikkei
newswire; Figure \ref{fig:corpusstats} gives some statistics on this
corpus.
Five 500-sequence {held-out} subsets were obtained from this
corpus for parameter training and algorithm testing; the rest of the
data served as the unsegmented corpus from which we derived character
$n$-gram counts.  The five subsets were extracted by randomly selecting
kanji sequences of at least ten characters in length -- recall from
our discussion above that long kanji sequences are the hardest to
segment \cite{Takeda+Fujisaki:87a}.  As can be seen from Figure
\ref{fig:testsets}, the average sequence length was around 12 kanji.
The subsets were then annotated following the segmentation policy outlined in
Section \ref{sec:annotate} below; as Figure \ref{fig:testsets} shows, the
average word length was between two and three kanji characters.
Finally, each of these five subsets was randomly split into a
50-sequence parameter-training set and a 450-sequence test set.  Any
sequences occurring in both a test set and its corresponding
parameter-training set were discarded from the parameter-training set
in order to guarantee that these sets were disjoint (recall that we
are especially interested in the unknown word problem); typically, no more
than five sequences needed to be removed.

\begin{figure}
\begin{center}
\begin{tabular}{lccc}
     &  average sequence length &  number of words  &    average word length    \\
test1 &    12\hpt27      &      2554    &       2\hpt40  \\  
test2      &     12\hpt40      &      2574    &       2\hpt41  \\    
test3      &     12\hpt05      &      2508    &       2\hpt40  \\   
test4      &     12\hpt12      &      2540    &       2\hpt39  \\   
test5      &     12\hpt22      &      2556    &       2\hpt39  \\ 
\end{tabular}
\end{center}
\caption{\label{fig:testsets} The statistics of the five held-out datasets.
Lengths are in characters.}
\end{figure}

Parameter training for our algorithm consisted of trying all nonempty
subsets of $\set{2,3,4,5,6}$ for the\ours\  set of $n$-gram orders $\nSet$
and all values in $\set{.05,.1,.15, \ldots, 1}$ for the threshold
$t$.  In cases where two parameter settings gave the same
training-set performance, ties were deterministically broken by
preferring smaller cardinalities of $\nSet$, shorter $n$-grams in
$\nSet$, and larger threshold values, in that order.  

The results for Chasen and Juman reflect the lexicon additions
described in Section \ref{sec:baselines}.

\subsection{Annotation Policies for the Held-Out Sets}
\label{sec:annotate}

To obtain the gold-standard annotations, the five held-out sets were
segmented by hand.  This section describes the policies we followed to
create the segmentations.  Section \ref{sec:annotate-general} gives an
overview of the basic guidelines we used.  Section
\ref{sec:annotate-details} goes into further detail regarding some
specific situations in Japanese.

\subsubsection{General policy}
\label{sec:annotate-general}

The motivation behind our annotation policy is Takeda and Fujisaki's
\shortcite{Takeda+Fujisaki:87a} observation that many kanji compound
words consist of two-character {\em stem} words together with
one-character prefixes and suffixes.  To handle the question of
whether affixes should be treated as separate units or not, we devised
a two-level bracketing scheme, so that accuracy can be measured with
respect to either level of granularity without requiring
re-annotation.  At the {\em word level}, stems and their affixes are
bracketed together as a single unit.  At the {\em morpheme level},
stems are divided from their affixes.  For example,
the sequence
\newcommand{\kheight}{.3cm}
\psfig{figure=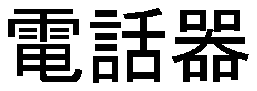,height=\kheight}
(\gloss{telephone}) would have the segmentation
\psfig{figure=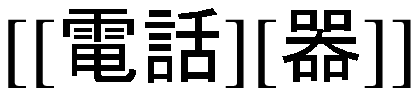,height=\kheight}  (\gloss{[[phone][device]]}) because
\psfig{figure=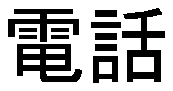,height=\kheight} (\gloss{phone}) can
occur on its own, but \psfig{figure=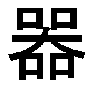,height=\kheight}
(\gloss{-device}) can appear only as an affix.  

We observe that both the word level and the morpheme level are
important in their own right.  Loosely speaking, word-level segments
correspond to discourse entities; they appear to be a natural unit for
native Japanese speakers (see next paragraph), and seem to be the
right granularity for document indexing, question answering, and other
end applications.  On the other hand, morpheme-level brackets
correspond to strings that cannot be further divided without loss of
meaning --- for instance, if one segments
\psfig{figure=denwa.eps,height=\kheight} into
\psfig{figure=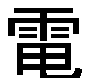,height=\kheight} (\gloss{electricity})
and \psfig{figure=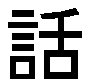,height=\kheight} (\gloss{speech}),
the meaning of the phrase becomes quite different.  Wu
\shortcite{Wu:98a} argues that indivisible units are the proper output
of segmentation algorithms; his point is that segmentation should be
viewed as a pre-processing step that can be used for many purposes.

Three native Japanese speakers participated in the annotation process: one
(the first author) segmented all the held-out data based on the above
rules, and the other two reviewed\footnote{We did not ask the other
two to re-annotate the data from scratch because of the tedium of the
segmentation task.}   350 sequences in total.  
The percentage of agreement with the first person's bracketing was
98.42\%: only 62 out of 3927 locations were contested by a
verifier.  Interestingly, all disagreement was at the morpheme level.

\subsubsection{Special situations}
\label{sec:annotate-details}

It sometimes happens that a character that can occur on its own serves
as an affix in a particular case.  Our decision was to annotate such
characters based on their role.  For example, although both
\psfig{figure=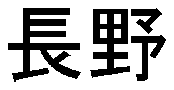,height=\kheight} (\gloss{Nagano})
and \psfig{figure=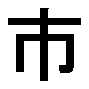,height=\kheight} (\gloss{city}) can
appear as individual words,
\psfig{figure=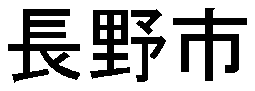,height=\kheight} (\gloss{the city
of Nagano}) is bracketed as
\psfig{figure=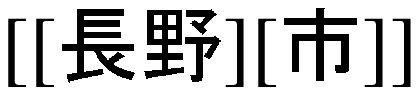,height=\kheight}, since here
\psfig{figure=shi.eps,height=\kheight} serves as a suffix.  As
a larger example, here is the annotation of a full sequence appearing
in our datasets: \\ \epsfscaledbox{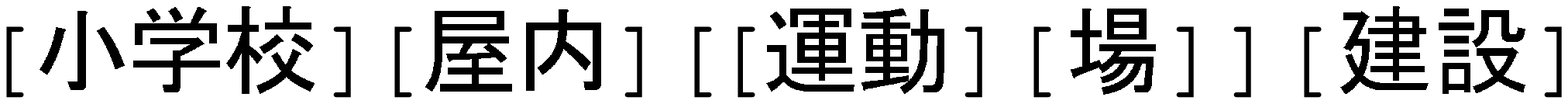}{3.2in}
(\gloss{[elementary school][building interior][[sports][area]][construction]}, i.e.  \gloss{construction of an elementary-school
indoor arena}).  Each bracket encloses a noun. Here,
\psfig{figure=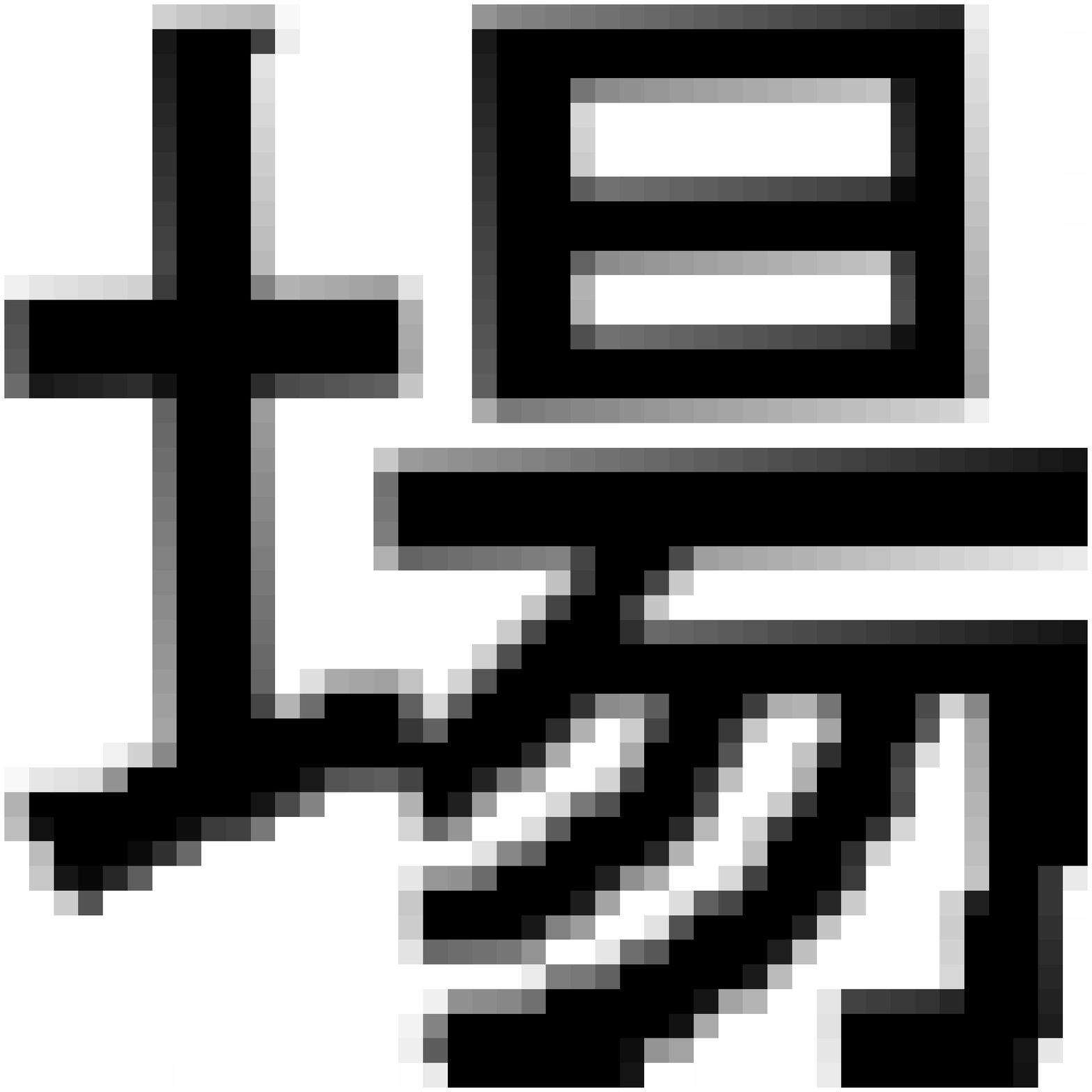,height=\kheight} is treated as an
affix and thus bracketed as a morpheme, although it can also appear as
an independent word as well.  On the other hand,
\psfig{figure=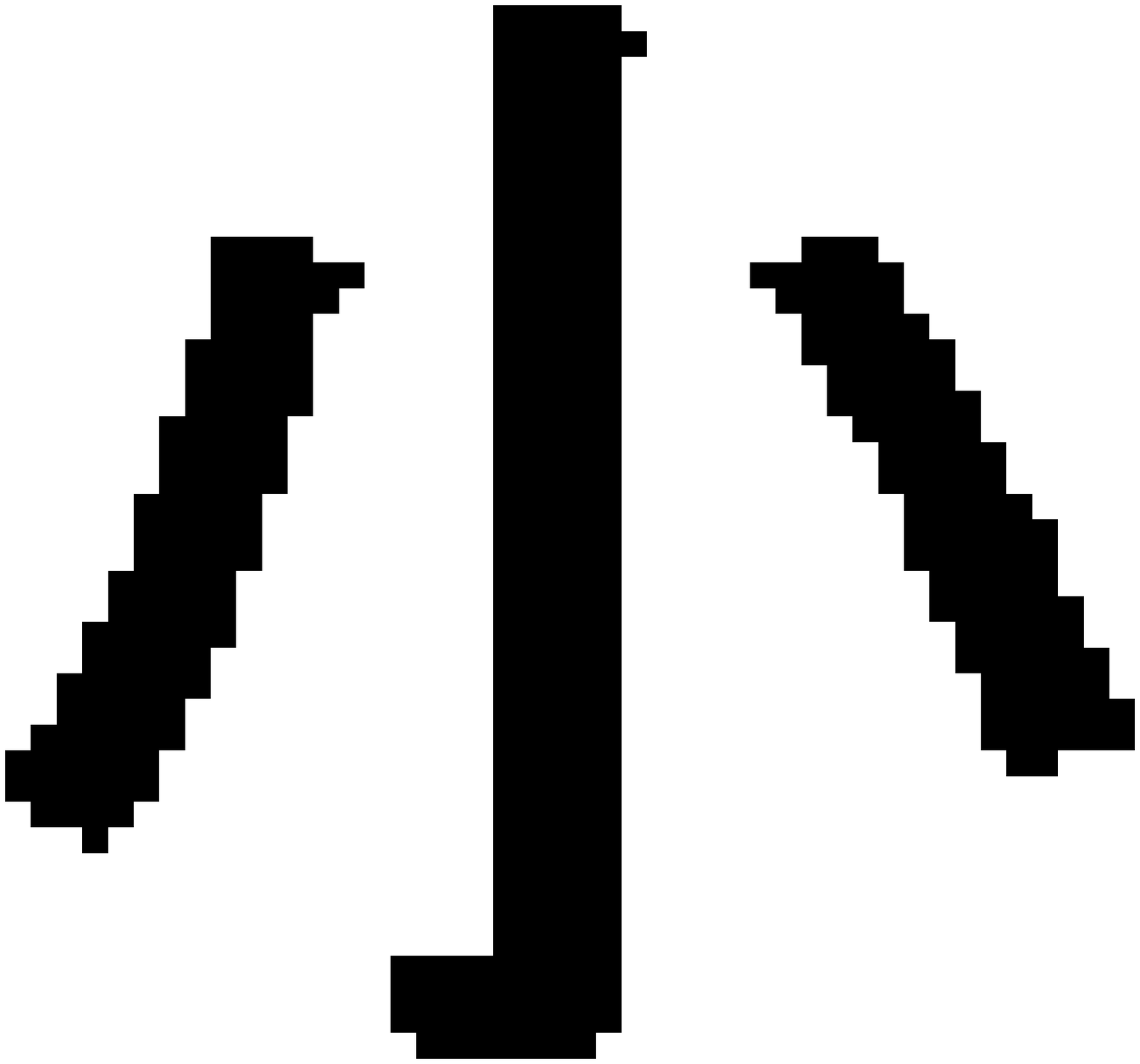,height=\kheight} has not been placed in
its own segment even though it is by itself also a word (\gloss{small}) -- otherwise, we
would get \gloss{small school} rather than the intended
\gloss{elementary school}.

Numeric strings (i.e., sequences containing characters denoting
digits) such as \psfig{figure=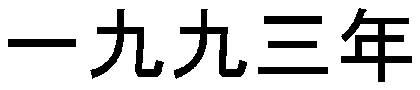,height=\kheight}
(\gloss{one-nine-nine-three-year}, \gloss{the year 1993}), are another
grey area.  In fact, Chasen and Juman appear to implement different
segmentation policies with respect to this type of character
sequence.  The issue arises not just from the digit characters
(should each one be treated as an individual segment, or should they
be aggregated together?), but also {\em classifiers}. In Japanese,
classifiers are required after numbers to indicate the unit or type
of the quantity.  That is, one must say
\psfig{figure=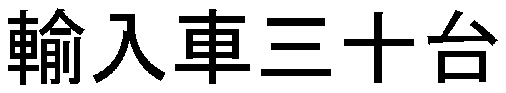,height=\kheight}
(\gloss{transfer\footnote{This word is two kanji characters long.}-car-three-ten-$\langle$car-unit$\rangle$},
\gloss{thirty imported cars}); omitting the
classifier so as to literally say ``thirty imported cars''
(*\psfig{figure=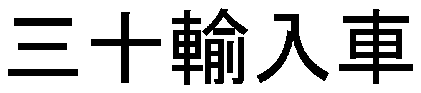,height=\kheight}) is not
permissible.  

Our two-level bracketing scheme makes the situation relatively easy to
handle.  In our annotations, the word level forms a single unit out of
the digit characters and subsequent classifier.  At the morpheme
level, the characters representing the actual number are treated as a
single unit, with the classifier serving as suffix.  For instance,
\psfig{figure=1993nen.eps,height=\kheight} is segmented as
\psfig{figure=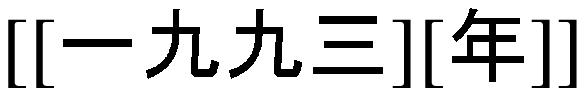,height=\kheight}
(\gloss{[[one-nine-nine-three][year]]}).
Also, to even out the inconsistencies of the morphological analyzers
in their treatment of numerics, we altered the output of all the
algorithms tested, including our own, so as to conform to our
bracketing; hence, our numerics segmentation policy did not affect the
relative performance of the various methods.  Since in practice it is
easy to implement a post-processing step that concatenates together
number kanji, we do not feel that this change misrepresents the
performance of any of the algorithms.  Indeed, in our experience the
morphological analyzers tended to over-segment numeric sequences, so this
policy is in some sense biased in their favor.

\subsection{Evaluation Metrics}
\label{sec:eval-measures}

\begin{figure*}
\begin{center}
\begin{tabular}{lcccc} \hline
   & Word errors & Morpheme errors    &
\multicolumn{2}{c}{Compatible-bracket errors:} \\ 
& (prec.,recall) &  (prec.,recall)   & Crossing  & Morpheme-dividing \\ \hline \hline
{\bf {\tt [[data][base]][system]}} & - & -& -&  -  \\
     {\tt |database|system|}  & 0,0 &  1,2 & 0 & 0 \\
     {\tt |data|base|system|}  & 2,1 &  0,0 & 0 & 0 \\
     {\tt |data|basesystem|} & 2,2 & 1,2 & 1 & 0 \\
     {\tt |database|sys|tem|} & 2,1 & 3,3 &  0 & 2 \\ \hline
\end{tabular}
\end{center}
\caption{\label{fig:erroregs} Examples illustrating segmentation
errors.  The first line gives the annotation segmentation: because
``data base'' and ``database'' are interchangeable, we have the
segmentation ``[[data][base]]''.  The word and morpheme columns
enumerate both precision and recall errors.  For example, the last
line commits two word precision errors because neither ``sys'' nor
``tem'' are annotated as words. The one word recall error comes from
missing ``system''.}
\end{figure*}

We used a variety of evaluation metrics in our experiments.  These
include the standard metrics of word precision, recall, and F measure;
and morpheme precision, recall, and F measure.  Furthermore, we
developed two novel metrics, the {\em compatible brackets} and {\em
all-compatible brackets} rates, which combine both levels of
annotation brackets.

\subsubsection{Word and morpheme precision, recall, and F}
 Precision and recall are
natural metrics for word segmentation.  Treating a proposed
segmentation as a non-nested bracketing (e.g.,
``$\vert$AB$\vert$C$\vert$'' corresponds to the bracketing
``[AB][C]''), {\em word precision} ($P$) is defined as the percentage
of proposed brackets that exactly match word-level brackets in the
annotation; {\em word recall} ($R$) is the percentage of word-level
annotation brackets that are proposed by the algorithm in question;
and {\em word F} combines precision and recall via their harmonic
mean: $F = 2PR/(P + R)$.  See  Figure
\ref{fig:erroregs} for some examples of word precision and recall errors.

The morpheme metrics are all defined analogously to their word
counterparts.  See Figure \ref{fig:erroregs} for examples of morpheme
precision and recall errors.

\subsubsection{Compatible-brackets and all-compatible brackets rates}

Word-level and morpheme-level accuracy are natural performance
metrics. However, they are clearly quite sensitive to the test
annotation, which is an issue if there are ambiguities in how a
sequence should be properly segmented. According to Section
\ref{sec:annotate}, there was 100\% agreement among our native
Japanese speakers on the word level of segmentation, but there was a
little disagreement at the morpheme level.  Also, the authors of Juman
and Chasen may have constructed their default dictionaries using
different notions of word and morpheme than the definitions we used in
annotating the data.  Furthermore, the unknown word problem leads to
some inconsistencies in the morphological analyzers' segmentations.
For instance, well-known university names are treated as single
segments by virtue of being in the default lexicon, whereas other
university names are divided into the name and the word
``university''.

We therefore developed two more robust metrics, adapted from measures
in the parsing literature, that penalize
proposed brackets that would be incorrect with respect to {\em any}
reasonable annotation.  These metrics account for two types of errors.
The first, a {\em crossing bracket}, is a proposed bracket that
overlaps but is not contained within an annotation bracket
\cite{Grishman+Macleod+Sterling:92a}.  Crossing brackets cannot
coexist with annotation brackets, and it is unlikely that another
human would create them.\footnote{The situation is different
for parsing, where attachment decisions are involved.}  The second
type of error, a {\em morpheme-dividing bracket}, subdivides a
morpheme-level annotation bracket; by definition, such a bracket
results in a loss of meaning.  See Figure \ref{fig:erroregs} for some
examples of these types of errors.

We define a {\em compatible bracket} as a proposed bracket that is
neither crossing nor morpheme-dividing.  The {\em compatible brackets
rate} is simply the compatible brackets precision, i.e., the percent
of brackets proposed by the algorithm that are compatible brackets.
Note that this metric accounts for different levels of segmentation
simultaneously.

We also define the {\em all-compatible brackets rate}, which is the
fraction of sequences for which {\em all} the proposed brackets are
compatible.  Intuitively, this function measures the ease with which a
human could correct the output of the segmentation algorithm: if the
all-compatible brackets rate is high, then the errors are concentrated
in relatively few sequences; if it is low, then a human doing
post-processing would have to correct many sequences.

It is important to note that neither of the metrics based on
compatible brackets should be
directly optimized in training, because an algorithm that simply brackets an
entire sequence as a single word will be given a perfect score.
Instead, we apply it only to measure the quality of test results
derived by training to optimize some other criterion.  Also, for the
same reason, we only compare brackets rates for algorithms
with reasonably close precision, recall, and/or F measure.  Thus, the
compatible brackets rates and all-compatible brackets rates should be
viewed as auxiliary goodness metrics.

\section{Experimental Results}
\label{sec:results}

We now report average results for all the methods over
the five test sets using the optimal parameter settings for the
corresponding training sets.  In all performance graphs, the ``error
bars'' represent one standard deviation.  

The section is organized as follows.  We first present 
our main experiment, reporting
the results of
comparison of \ours\ to Chasen and Juman (Section
\ref{sec:cf-morph}).  We see that our mostly-unsupervised algorithm is
competitive with and sometimes surpasses their performance.  
Since these experiments demonstrate the promise of using a purely
statistical approach, 
Section
\ref{sec:cf-sun} describes our comparison against another
mostly-unsupervised algorithm, \sst\ \cite{Sun+Shen+Tsou:98a}.
These experiments show that
our method, which is based on very simple statistics, can
substantially outperform methods based on more sophisticated functions
such as mutual information.
Finally, Section
\ref{sec:analysis} goes into further discussion and analysis.

\subsection{Comparison Against Morphological Analyzers}
\label{sec:cf-morph}

\subsubsection{Word and morpheme accuracy results} 
To produce word-level segmentations from Juman and Chasen, we altered
their output by concatenating stems with their affixes, as
identified by the part-of-speech information the analyzers provided.

\begin{figure}
\epsfscaledbox{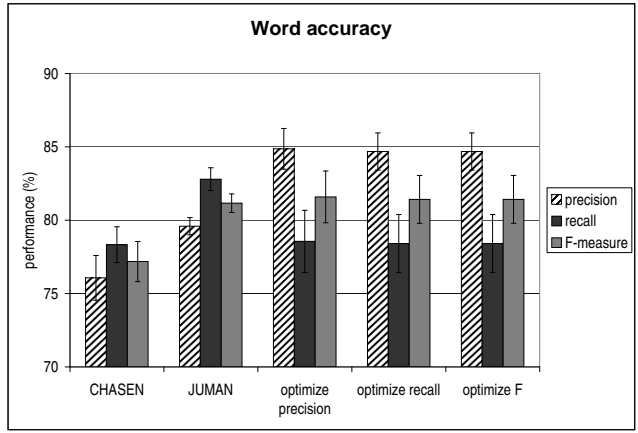}{3.8in}
\caption{\label{fig:word} Word accuracy, Chasen and Juman vs. \ours.
The three rightmost groups show \ours's performance when parameters
were tuned for different optimization criteria.}
\end{figure}

\begin{figure}
\begin{center}
\epsfscaledbox{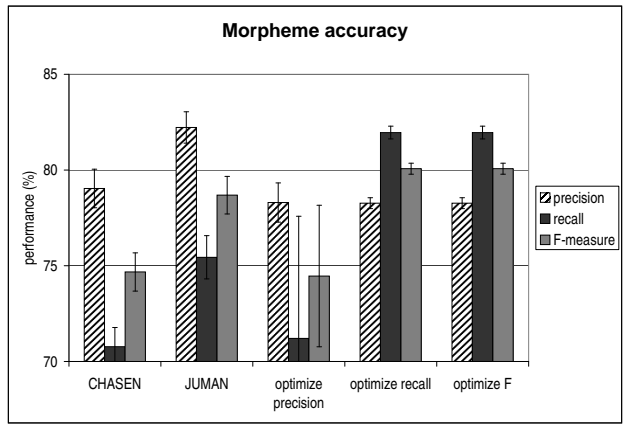}{3.6in}
\caption{\label{fig:morph_data} Morpheme accuracy, Chasen and Juman
vs. \ours.
}  
\end{center}
\vspace*{-.4in}
\end{figure}

Figure \ref{fig:word} shows word accuracy for Chasen, Juman, and our
algorithm for parameter settings optimizing word precision, recall,
and F-measure rates.  As can be seen, in this case the choice of
optimization criterion did not matter very much.  \ours\
achieves more than five percentage-points higher precision than Juman
and Chasen.  \ours's recall performance is also respectable in comparison
to the morphological analyzers, falling (barely) between that of Juman
and that of Chasen.  Combined performance, as recorded by the F-measure,
was about the same as Juman, and 4.22 percentage points higher than
Chasen.

We also measured {\em morpheme precision, recall}, and {\em F
measure}, all defined analogously to their word counterparts.  Figure \ref{fig:morph_data} shows the results.  Unlike
in the word accuracy case, the choice of training criterion clearly
affected performance.

We see that \ours's morpheme precision was slightly lower than that of
Chasen, and noticeably worse than for Juman.  However, when we optimize for
precision, the results for all three metrics are on average quite
close to Chasen's.  \ours's recall results are substantially
higher than for the morphological analyzers when recall was its
optimization criterion.  In terms of combined performance, tuning our
algorithm for the F-measure achieves higher F-measure performance than
Juman and Chasen, with the performance gain due to enhanced recall.

Thus, overall, we see that our method yields results rivalling those
of morphological analyzers.

\subsubsection{Compatible brackets results}

\begin{figure}
\epsfscaledbox{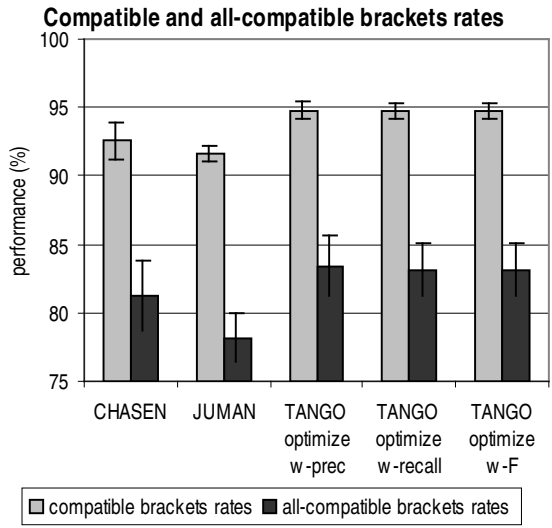}{3in}
\caption{\label{fig:xbrates} Compatible and all-compatible brackets rates,  word-accuracy training.}  
\end{figure}

\begin{figure}
\epsfscaledbox{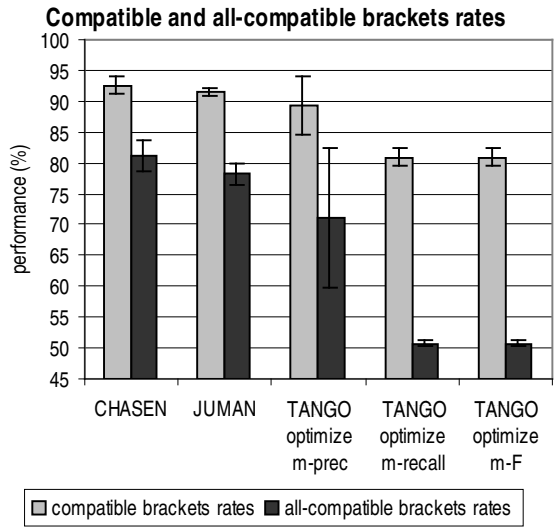}{3in}
\caption{\label{fig:xbrates_morph} Compatible and all-compatible brackets rates, morpheme-accuracy training.}
\end{figure}

Figure \ref{fig:xbrates} depicts the compatible brackets and
all-compatible brackets rates for Chasen, Juman, and \ours, using the
three word accuracy metrics as training criteria for our method.
\ours\  does better on the compatible brackets rate and the
all-compatible brackets rates than the morphological analyzers do,
regardless of training optimization function used.

The results degrade if morpheme accuracies are used as training
criteria, as shown in Figure \ref{fig:xbrates_morph}.  Analysis of the
results shows that most of the errors come from morpheme-dividing
brackets: the parameters optimizing morpheme recall and F-measure are
$\nSet = \set{2}$ and $t=.5$, which leads to a very profligate policy
in assigning boundaries.

\subsubsection{Discussion}

Overall, we see that our algorithm performs very well with respect to
the morphological analyzers when word accuracies are used for
training.  At the morpheme level, the relative results for our algorithm
in comparison to Chasen and Juman are mixed; recall and F are
generally higher, but at the price of lower compatible brackets rates.
The issue appears to be that single-character segments are common at
the morpheme level because of affixes. The only way our algorithm can
create such segments is by the threshold value, but too low a
threshold causes precision to drop.  

A manual (and quite tedious) analysis of the types of errors produced
in the first test set reveals that the major errors our algorithm
makes are mis-associating affixes (e.g., wrongly segmenting
\psfig{figure=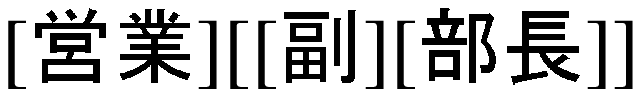,height=\kheight} as
\psfig{figure=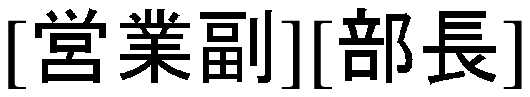,height=\kheight}) and
fusing single-character words to their neighbors (e.g. wrongly segmenting
\psfig{figure=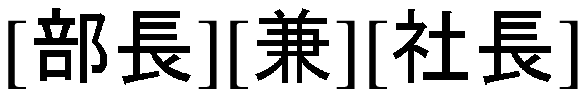,height=\kheight} as 
\psfig{figure=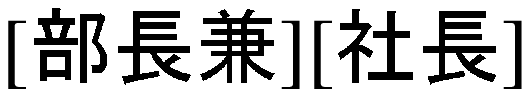,height=\kheight}).  Both
of these are mistakes involving single characters, a problem just
discussed.  Another common error was the incorrect concatenation of
family and given names, but this does not seem to be a particularly
harmful error.  On the other hand, the morphological analyzers made
many errors in segmenting proper nouns, which are objects of prime
importance in information extraction and related tasks, whereas our
algorithm made far fewer of these mistakes.

In general, we conclude that \ours\ produces high-quality
word-level segmentations rivaling and sometimes outperforming
morphological analyzers, and can yield morpheme accuracy results
comparable to Chasen, although the bracket-metrics quality of the
resultant morpheme-level segmentations is not as good. The algorithm
does not fall prey to some of the problems in handling proper nouns
that lexico-syntactic approaches suffer. Developing methods for
improved handling of single-character morphemes would enhance the
method, and is a direction for future work.

\subsection{Comparison against \sst}
\label{sec:cf-sun}

We further evaluated our algorithm by comparing it against another
mostly-unsuper\-vised algorithm: that of Sun, Shen, and Tsou
\shortcite{Sun+Shen+Tsou:98a} (henceforth \sst), which, to our
knowledge, is the segmentation technique most similar to the one we
have proposed.  Our goal in these experiments was to  investigate the
null hypothesis that any reasonable $n$-gram-based method could yield
results comparable to Chasen and Juman.

\sst\ also performs segmentation without using a grammar, lexicon, or
large amounts of annotated training data; instead, like \ours, it
relies on statistics garnered from an unsegmented corpus.  Although
Sun et al. implemented their method on Chinese (which also lacks space
delimiters) instead of Japanese, they designed \sst\ to be able to
account for ``two common construction types in Chinese word formation:
`2 characters + 1 character' and `1 character + 2 characters''
(pg. 1269).  As pointed out in Section \ref{sec:annotate-general}
above, Japanese kanji compound words often have a similar structure
\cite{Takeda+Fujisaki:87a}; hence, we can apply \sst\ to kanji as
well, after appropriate parameter training.

\subsubsection{The \sst\ Algorithm}
\label{sec:sun}

Like our algorithm, \sst\ seeks to determine whether a given location
constitutes a word boundary by looking for evidence of cohesion (or
lack thereof) between the characters bordering the location.  Recall
the situation described by Figure \ref{fig:intuition}: we have the
character sequence ``A B C D W X Y Z'', and must decide whether the
location between D and W should be a boundary.  The evidence \sst\
uses comes from two sources.  The first is the (pointwise or Fano)
{\em mutual information}\footnote{This function (and its more
probabilistically-motivated cousin the {\em Shannon} or {\em average
mutual information} \cite{Cover+Thomas:91a}) has a long and
distinguished history in natural language processing. In particular,
it has been used in other segmentation algorithms
\cite{Sproat+Shih:90a,Wu+Su:93a,Lua:95a}; interestingly, Magerman
and Marcus \shortcite{Magerman+Marcus:90a} also used a variant of the
mutual information to find constituent boundaries in English.}
between D and W: $mi(\D,\W) = \log(p(\D,\W)/(p(\D)p(\W)))\,.$ The
other statistic is the difference in {\em t-score}
\cite{Church+Hanks:90a}  for
the two trigrams CDW and DWX straddling the proposed boundary:
$$dts(\D,\W)  = \frac{p(\W|\D) - P(\D|\C)}{\sqrt{{\rm
var}(p(\W|\D)) + {\rm var}(p(\D|\C))}} -
\frac{p(\X|\W) - P(\W|\D)}{\sqrt{{\rm
var}(p(\X|\W)) + {\rm var}(p(\W|\D))}} 
$$
Note that both the mutual information and the difference in t-score
can be computed from character bigram statistics alone, as gathered
from an unsegmented corpus.

Segmentation decisions are made based on a threshold $\theta$ on the
mutual information, and on thresholds for the first and secondary local
maxima and minima of the difference in t-score.  There are six such
free {\em extremum parameters}. In addition, we treated $\theta$ as
another free parameter in our re-implementation; Sun et al. simply
set it to the same value (namely, 2.5) that Sproat and Shih
\shortcite{Sproat+Shih:90a}) used.  These seven parameters together
yield a rather large parameter search space.  For our implementation
of the parameter training phase (Sun et al.'s paper
does not give details), we set each of the
six extremum thresholds\footnote{The values to be thresholded are
defined to be non-negative.} to all the values in the set $\set{0, 50,
100, 150, 200}$ and the mutual information threshold\footnote{It is
the case that the pointwise mutual information can take on a negative
value, but this indicates that the two characters occur together less
often than would be expected under independence assumptions. Therefore,
since the threshold is used to determine locations that should {\em
not} be boundaries, non-negative thresholds suffice.
} 
to all values in the set $\set{0, 1.25, 2.5, 3.75, 5}$.  Ties in
training set performance --- which were plentiful in our experiments
--- were broken by choosing smaller parameter values.  We used the same
count tables (i.e. with singleton $n$-grams deleted and unseen
$n$-grams given a count of one) as for \ours.

While local extrema and thresholds also play a important role in our
method, we note that \sst\ is considerably more complex 
than the algorithm we have proposed.  
Also observe that it makes use
of conditional probabilities, whereas we simply compare frequencies.

\subsubsection{\sst\ Results}

\begin{figure}
\epsfscaledbox{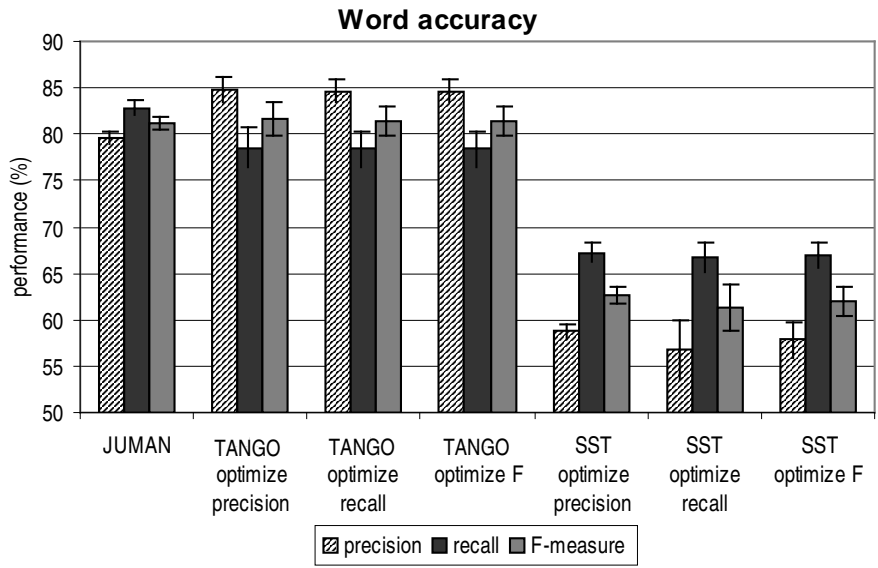}{4.8in}
\caption{\label{fig:sun-word} Word accuracy, Juman vs. \ours\
vs. \sst.}
\end{figure}

\begin{figure}
\epsfscaledbox{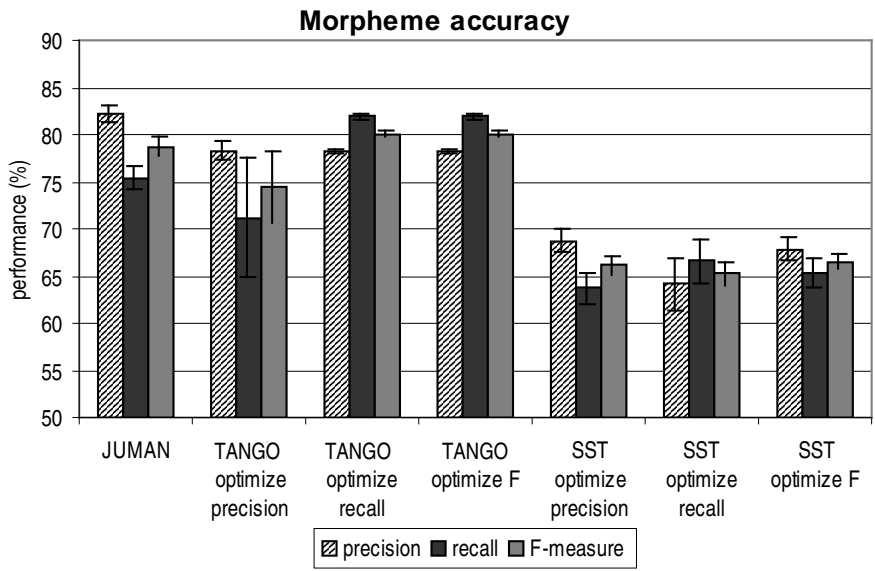}{4.8in}
\caption{\label{fig:sun-morph} Morpheme accuracy, Juman vs. \ours\
vs. \sst.}
\end{figure}

Figures \ref{fig:sun-word} and \ref{fig:sun-morph} show the word and
morpheme accuracy results.\footnote{Church and Hanks
\shortcite{Church+Hanks:90a} actually suggested using two different
ways to estimate the probabilities used in the t-score: the {\em
maximum likelihood estimate} (MLE) and the {\em expected likelihood
estimate} (ELE).  According to Sun (personal communication), the MLE
was used in the original implementation, although Church and Hanks
suggested that the ELE would be better.  We experimented with both,
finding that they gave generally similar results.  For consistency
with Sun et al.'s experiments, we plot the MLE results only. The only
case in which using the ELE rather than
the MLE made a substantial difference
was in morpheme precision, which improved by five percentage points
(not enough to change the relative performance rankings).
} 
Clearly, \sst\ does not perform as well as 
Juman
at either level of segmentation granularity.  Since the
accuracy results are so much lower than the other algorithms,  we do
not apply the brackets metrics, for the reasons discussed above.

It is clear that our algorithm outperforms \sst, even
though \sst\ uses much more sophisticated statistics. Note that we
incorporate $n$-grams of different orders, whereas \sst\ essentially
relies only on character bigrams, so at least some of the gain may be
due to our using more sources of information from the unsegmented
training corpus.  We do not regard this as an unfair advantage because
incorporating higher-order $n$-grams is easy to do for our method, and
adds no conceptual overhead, whereas extending the mutual information
to trigrams and larger $n$-grams can be a complex procedure
\cite{Magerman+Marcus:90a}.

Stepping back, though, we note that our goal is not to emphasize the
comparison with the particular algorithm \sst,
especially since it was developed for Chinese rather than Japanese.
Rather, what we conclude from these experiments is that it is not
necessarily easy for weakly-supervised statistical algorithms to do as
well as morphological analyzers, even if they are given large amounts of
unsegmented data as input and 
even if they employ well-known
statistical functions.
This finding further validates the \ours\
algorithm.

\subsection{Further Explorations}
\label{sec:analysis}

Given \ours's good performance in comparison to Chasen and Juman, we
examined the role of the amount of parameter-training data \ours\ is
given as input,  and investigated the contributions of the algorithm's
two parameters.

\subsubsection{Minimal human effort is needed.} In contrast to our
mostly-unsupervised method, morphological analyzers need a lexicon and
grammar rules built using human expertise.  The workload in creating
dictionaries on the order of hundreds of thousands of words (the size
of Chasen's and Juman's default lexicons) is clearly much larger than
annotating the small parameter-training sets for our algorithm.  \ours\
also obviates the need to segment a large amount of parameter-training
data because our algorithm draws almost all its information from an
unsegmented corpus.
Indeed, the only human effort involved in our algorithm is
pre-segmenting the five 50-sequence parameter training sets, which
took only 42 minutes.\footnote{Annotating all the {\em test} data for
running comparison experiments takes considerably longer.}  In
contrast, previously proposed supervised approaches to Japanese
segmentation have used annotated training sets ranging from 1000-5000
sentences \cite{Kashioka+al:98a} to 190,000 sentences
\cite{Nagata:96a}.

\begin{figure*}
\begin{center}
\fbox{\begin{tabular}{lrrrrr}
          & \ours\ (50) & \ours\ (5)             & Juman (5) & \ours\
          (5)         &   \ours\ (5)  \\   
          &  vs. &  vs. 		   &  vs.        &    vs.	      &  vs. \\
          & Juman (50) &  Juman (5) 	   &  Juman (50) & \ours\ (50)     & Juman (50) \\
\hline	 				                
precision  &   +5\hpt27      & +6\hpt18  & -1\hpt04      & -0\hpt13     & +5\hpt14 \\
recall     &  -4\hpt39       & -3\hpt73  & -0\hpt63      & +0\hpt03     & -4\hpt36\\
F-measure &   +0\hpt26       & +1\hpt14  & -0\hpt84      & +0\hpt04     & +0\hpt30\\
\end{tabular}
} 
\caption{\label{fig:word_data_small} Average relative word accuracy as a function of training set size.  
``5'' and ``50'' denote  training set size
 before discarding overlaps with the test sets.}  
\end{center}
\end{figure*}

To test how much annotated training data is actually necessary, we
experimented with using miniscule parameter-training sets: five sets
of only {\em five} strings each (from which any sequences repeated in
the test data were discarded).  It took only 4 minutes to perform the
hand segmentation in this case.  

Figure \ref{fig:word_data_small} shows the results. For instance, we
see that training \ours\ on five training sequences yielded 6.18
percentage points higher precision than using Juman with the same five
training sequences; thus, our algorithm is making very efficient use of
even very small training sets.  The third data column proves that the
information we provide Juman from the annotated training data does
help, in that using smaller training sets results in lower
performance.  According to the fourth data column, performance doesn't
change very much if our algorithm gets only five training sequences
instead of 50.  Interestingly, the last column shows that if our
algorithm has access to only five annotated sequences when Juman has
access to ten times as many, we still achieve better precision
and comparable F-measure.

\subsubsection{Both the local maximum and threshold conditions contribute.}
Recall that in our algorithm, a location $k$ is deemed a word boundary
if the evidence $v_\nSet(k)$ is either (1) a local maximum or (2) at
least as big as the threshold $t$.  It is natural to ask whether we
really need two conditions, or whether just one would
suffice.\footnote{Goldsmith \shortcite{Goldsmith:01a} notes that the
``peaks vs. threshold'' question (for different
functions) has arisen previously in the literature on morphology
induction for European languages; see page 158 of that paper for a short summary.} For
instance, while $t$ is necessary for producing single-character
segments, perhaps it also causes many recall errors, so that
performance is actually better when the threshold criterion is
ignored.

\newcommand{\M}{{\sf M}}
\newcommand{\T}{{\sf T}}

\begin{figure}
\epsfscaledbox{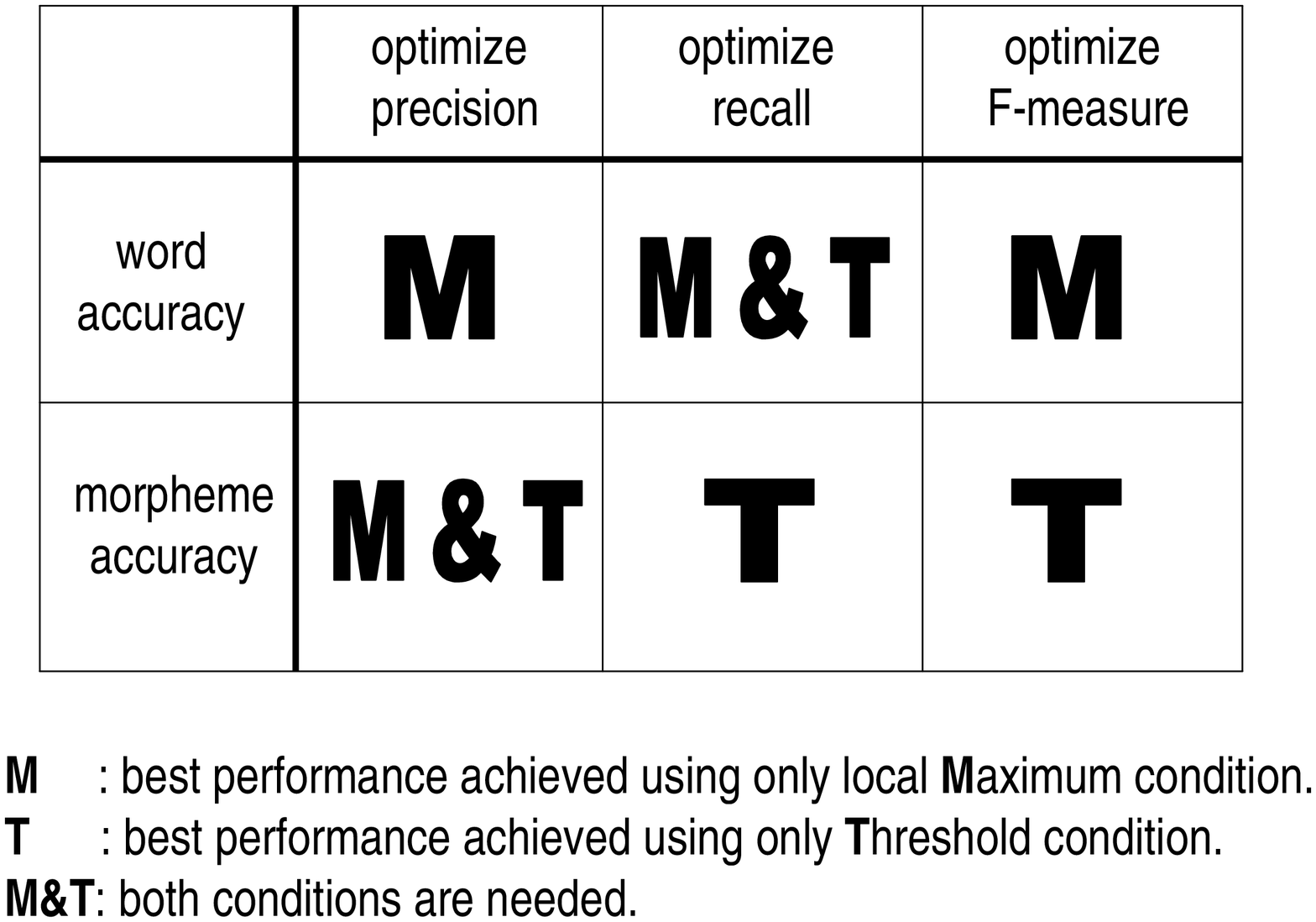}{3.5in}
\caption{\label{fig:lmax_threshold} Entries indicate whether best performance is
achieved using the local maximum condition (M), the threshold
condition (T), or both.}
\end{figure}

We therefore studied whether optimal performance could be achieved
using only one of the conditions. Figure \ref{fig:lmax_threshold}
shows that in fact both  contribute to producing good
segmentations.  Indeed, in some cases, both are needed to achieve the
best performance; also, each condition when used in isolation yields
suboptimal performance with respect to some performance metrics.  

\section{Related Work}
\label{sec:related}

\subsection{Japanese Segmentation} Many previously proposed segmentation methods
for Japanese text make use of either a pre-existing lexicon
\cite{Yamron+al:93a,Matsumoto+Nagao:94a,Takeuchi+Matsumoto:95a,Nagata:97a,Fuchi+Takagi:98a}
or a substantial amount of pre-segmented training data
\cite{Nagata:94a,Papageorgiou:94a,Nagata:96a,Kashioka+al:98a,Mori+Nagao:98a,Ogawa+Matsuda:99a}.
Other approaches bootstrap from an initial segmentation provided by a
baseline algorithm such as Juman
\cite{Matsukawa+Miller+Weischedel:93a,Yamamoto:96a}.

Unsupervised, non-lexicon-based methods for Japanese segmentation do
exist, but often have limited applicability when it comes to kanji.
Both Tomokiyo and Ries \shortcite{Tomokiyo+Ries:97a} and Teller and
Batchelder\shortcite{Teller+Batchelder:94a} explicitly avoid working
with kanji characters.  
Takeda and Fujisaki \shortcite{Takeda+Fujisaki:87a} propose the {\em
short unit model}, a type of Hidden Markov Model with
linguistically-determined topology, for segmenting certain types
of kanji compound words; in our five test datasets, we found that
13.56\% of the kanji sequences contain words that the short unit model
cannot handle.  

Ito and Kohda's \shortcite{Ito+Kohda:95a} completely unsupervised
statistical algorithm extracts high-frequency character $n$-grams as
pseudo-lexicon entries; a standard segmentation algorithm is then used
on the basis of the induced lexicon. But the segments so derived are
only evaluated in terms of perplexity and intentionally need not be
linguistically valid, since Ito and Kohda's interest is not in word
segmentation but in language modeling for speech.

\subsection{Chinese Segmentation} 

There are several varieties of statistical Chinese segmentation
algorithms that do not rely on syntactic information or a lexicon,
and hence could potentially be applied to Japanese.  One vein of work
relies on the mutual information \cite{Sproat+Shih:90a,Lua+Gan:94a};
\sst\ can be considered a
more sophisticated variant of these since it also incorporates
t-scores.  There are also algorithms that leverage pre-segmented
training data
\cite{Palmer:97a,Dai+Loh+Khoo:99a,Teahan+Web+McNab+Witten:00a,Brent+Tao:01a}.\footnote{Brent
and Tao's algorithm can, in principle, work with no annotated data
whatsoever.  However, their experiments show that at least 4096 words
of segmented training data were required to achieve precision over
60\%.}  Finally, the unsupervised, knowledge-free algorithms of Ge,
Pratt, and Smyth \shortcite{Ge+Pratt+Smyth:99a} and Peng and
Schuurmans \shortcite{Peng+Schuurmans:01a} both require use of the EM
algorithm to derive the most likely segmentation; in contrast, our
algorithm is both simple and fast.

\section{Conclusion}
\label{sec:conc}
In this paper, we have presented \ours, a simple, mostly-unsupervised
algorithm that segments Japanese sequences into words based on
statistics drawn from a large unsegmented corpus.
We evaluated performance on kanji with respect to several metrics, including
the novel compatible brackets and all-compatible brackets rates, and
found that our algorithm could yield performances rivaling that of
lexicon-based morphological analyzers.  Our experiments also showed
that a similar previously-proposed mostly-unsupervised algorithm could
not yield comparable results.

Since \ours\ learns effectively from unannotated data, it represents a
robust algorithm that is easily ported to other domains and
applications.  We view this research as a successful application of
the (mostly) unsupervised learning paradigm:  high-quality information
can be extracted from large amounts of raw data, sometimes by
relatively simple means.

On the other hand, we are by no means advocating a purely statistical
approach to the Japanese segmentation problem.  We have explicitly
focused on long sequences of kanji as being particularly problematic
for morphological analyzers such as Juman and Chasen, but these
analyzers are well-suited for other types of sequences.  We thus view
our method, and mostly-unsupervised algorithms in general, as
complementing knowledge-based techniques, and conjecture that hybrid
systems that rely on the strengths of both kinds of methods will prove
to be the most effective in the future.

\section*{Acknowledgments}  
We thank Michael Oakes, Hirofumi Yamamoto, and the anonymous referees
for very helpful comments.
We would like to acknowledge Takashi Ando and Minoru Shindoh for 
reviewing the annotations, and Sun Maosong for advice in implementing
the algorithm of Sun et al. \shortcite{Sun+Shen+Tsou:98a}.  Thanks to
Stuart Shieber for reference suggestions.  A preliminary version
of this work was published in the proceedings of NAACL 2001; we thank
the anonymous reviewers of that paper for their comments.  
This research was conducted while the first author was a graduate student
at Cornell University.
This
material is based on work supported in part by a grant from the GE
Foundation and by the National Science Foundation under ITR/IM grant
IIS-0081334.

\end{document}